%% file: main.tex
\documentclass{INTERSPEECH2023}

\interspeechcameraready

\usepackage{cite}
\usepackage{multirow}

\newcommand{\Sec}[1]{section~\ref{sec:#1}}
\newcommand{\Fig}[1]{Fig.~\ref{fig:#1}}
\newcommand{\Table}[1]{Table~\ref{tbl:#1}}

\newcommand{\method}[1]{\textsf{#1}}
\newcommand{\data}[1]{\textit{#1}}

\title{UniFLG: Unified Facial Landmark Generator from Text or Speech}
\name{Kentaro Mitsui, Yukiya Hono, Kei Sawada}
\address{rinna Co., Ltd., Tokyo, Japan}
\email{\{kemits,yuhono,keisawada\}@rinna.co.jp}

\begin{document}
\include{math}
\maketitle

\begin{abstract}
\vspace{-3pt}
Talking face generation has been extensively investigated owing to its wide applicability.
The two primary frameworks used for talking face generation comprise a text-driven framework, which generates synchronized speech and talking faces from text, and a speech-driven framework, which generates talking faces from speech.
To integrate these frameworks, this paper proposes a unified facial landmark generator (UniFLG).
The proposed system exploits end-to-end text-to-speech not only for synthesizing speech but also for extracting a series of latent representations that are common to text and speech, and feeds it to a landmark decoder to generate facial landmarks.
We demonstrate that our system achieves higher naturalness in both speech synthesis and facial landmark generation compared to the state-of-the-art text-driven method.
We further demonstrate that our system can generate facial landmarks from speech of speakers without facial video data or even speech data.

\end{abstract}

\vspace{-3pt}
\section{Introduction}
\vspace{-2pt}

\begin{figure}[t]
  \centering
  \includegraphics[width=\linewidth]{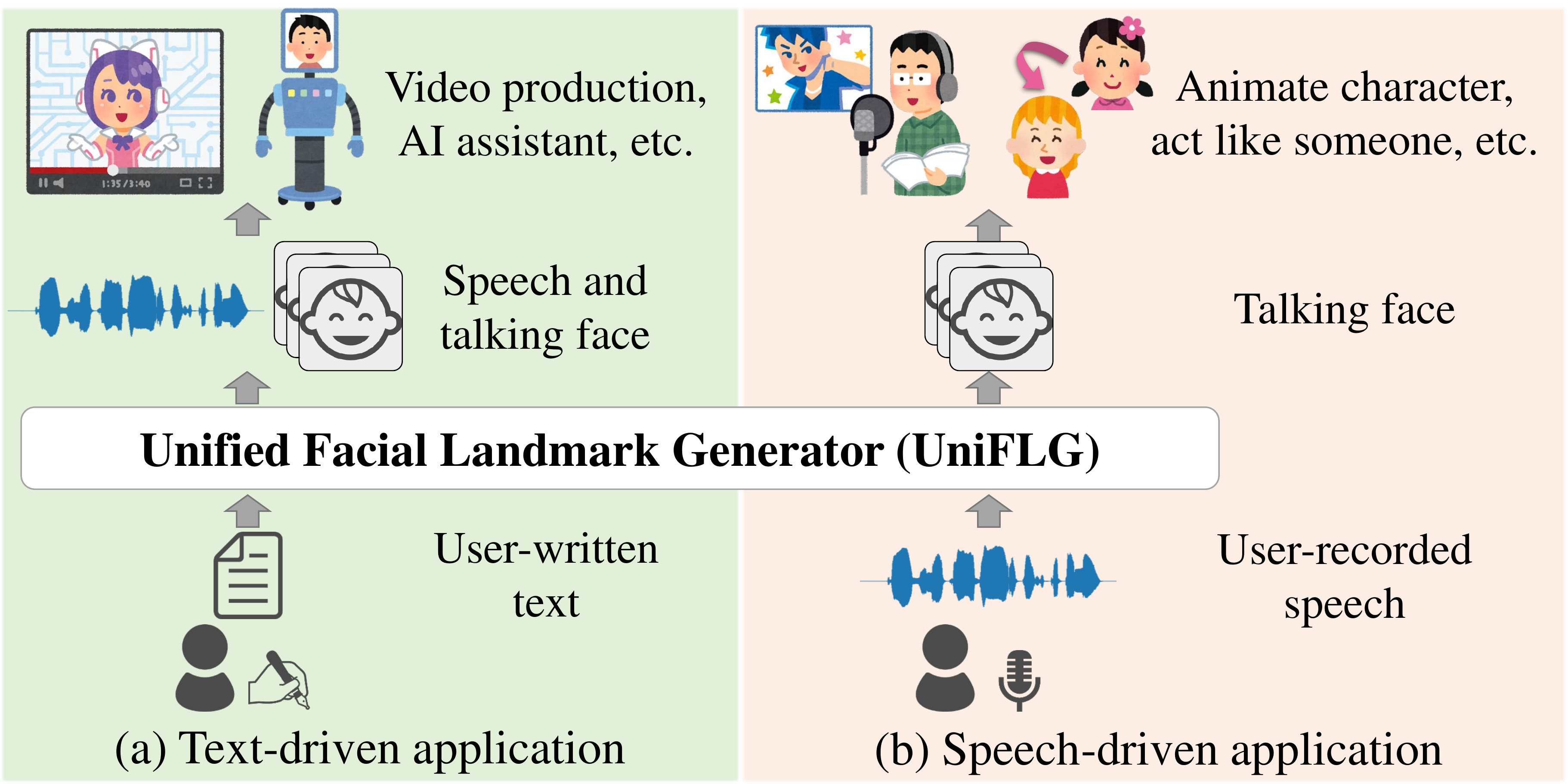}
  \caption{Major applications of talking face generation.}
  \label{fig:overview}
  \vspace{-5ex}
\end{figure}

In recent years, there has been growing interest in virtual humans and the metaverse, leading to an increased focus on the generation of natural talking faces~\cite{zhu2021deep}.
The applications of talking face generation can be broadly categorized into two groups, as depicted in \Fig{overview}.
The first group involves generating talking faces based on text inputs, which can be used for video production or multimodal chatbots~\cite{kumar2017obamanet,wang2021anyonenet,zhang2022text2video,li2021write,dahmani2019conditional,dahmani2021learning,abdelaziz2021avtacotron}.
In most cases, this group also requires simultaneous generation of speech synchronized with talking faces.
The second group involves generating talking faces synchronized with speech inputs, which can be used to animate characters' faces or to act like someone else~\cite{suwajanakorn2017obama,taylor2017deep,eskimez2018generating,zhou2020makeittalk,prajwal2020wav2lip,zhou2021pcavs,liang2022gcavt,zhang2022meta,deng2021unsupervised,ji2021audio}.
While this group usually requires removing the speaker identity to accommodate arbitrary speakers, it is also important to generate talking faces according to the speech emotions.
Although existing research has targeted only one of these groups, integrating them can produce a single versatile model for a variety of applications.
A current approach is to train a speech-to-face model for the second group of applications, and combine it with an external text-to-speech (TTS) model for the first group of applications~\cite{zhou2020makeittalk,wang2021anyonenet}.
However, this approach has several drawbacks; linguistic information cannot be utilized in face generation, the quality of generated talking faces is affected by the TTS quality, and additional inference time is required for TTS.

In this paper, we propose a unified facial landmark generator (UniFLG) to integrate text- and speech-driven talking face generation.
UniFLG has a TTS module based on the variational autoencoder (VAE)~\cite{kingma2014auto}, which makes it possible to acquire a time-aligned common representation of text and speech during TTS training.
Then, a landmark decoder, which is another module of UniFLG, generates facial landmarks from the intermediate representation.
This is beneficial for both text- and speech-driven generation;
during text-driven generation, speech and facial landmarks can be generated in parallel, resulting in faster inference and no error propagation from TTS.
During speech-driven generation, speaker identity is removed because the speech-based representation is learned to be shared with text-based representation.
To preserve speech emotions, we further introduce an utterance-level VAE to extract emotion embeddings and condition the landmark decoder on it.
Another important feature of this study is that we regard the facial landmark, a widely used low-dimensional representation of faces~\cite{kumar2017obamanet,wang2021anyonenet,zhang2022text2video,li2021write,dahmani2019conditional,dahmani2021learning,suwajanakorn2017obama,eskimez2018generating,zhou2020makeittalk,ji2021audio,yu2020multimodal,yu2021multimodal}, as common among speakers due to its little speaker dependence.
This assumption enables the landmark decoder to be trained with paired text, speech and facial videos of just one speaker, while the TTS module can be trained with existing multi-speaker corpora.

\vspace{-3pt}
\section{Related work}
\vspace{-2pt}

\begin{figure*}[t]
  \centering
  \includegraphics[width=\hsize]{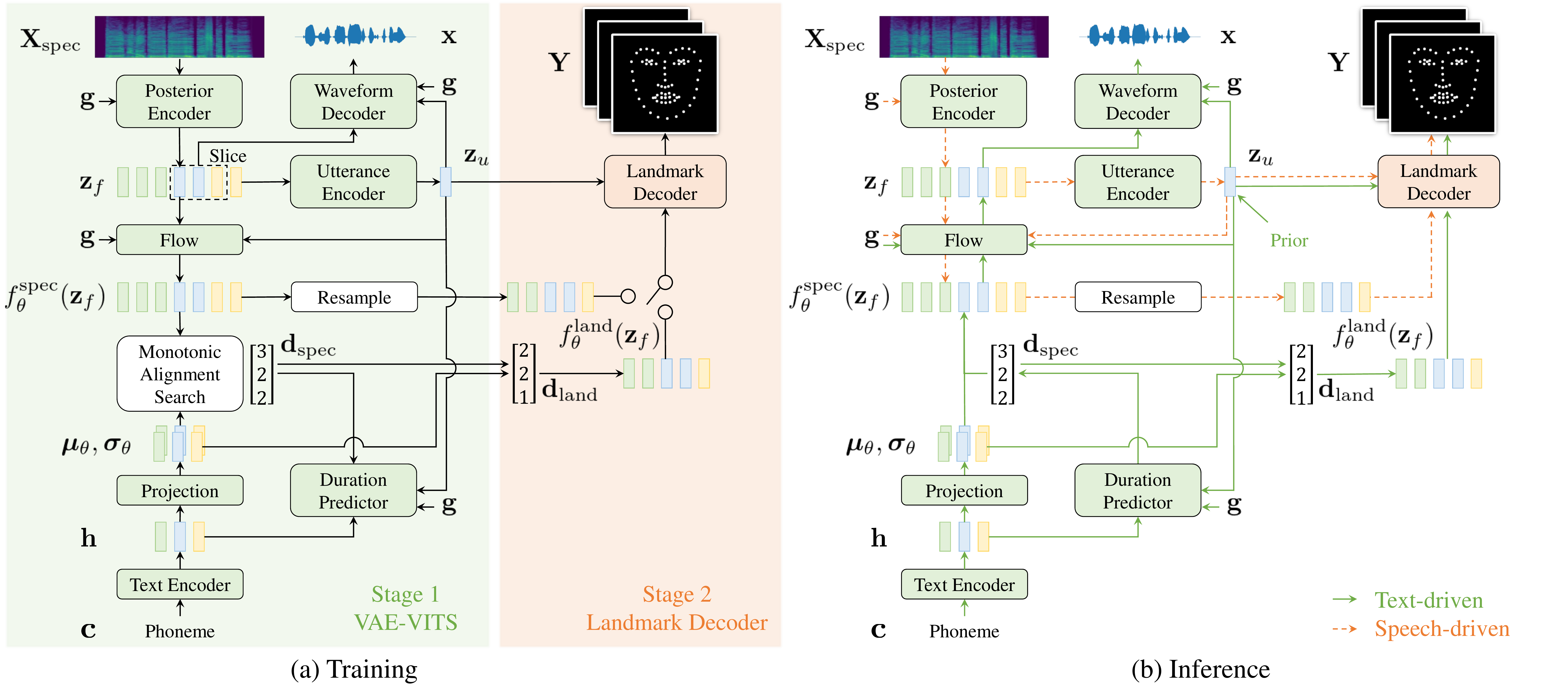}
  \caption{Conceptual diagram of the (a) training and (b) inference of UniFLG.}
  \label{fig:diagram}
\end{figure*}

\textbf{Text-driven talking face generation.}
Most of the text-driven talking face generation methods are accompanied by TTS.
Pipelined methods first synthesize speech from text and then generate talking faces in a speech-driven manner~\cite{kumar2017obamanet, wang2021anyonenet}, and text-based methods utilize temporal alignment between the text and speech obtained via TTS~\cite{zhang2022text2video, li2021write}.
On the other hand, audiovisual speech synthesis simultaneously generates speech and talking faces~\cite{dahmani2019conditional,dahmani2021learning,abdelaziz2021avtacotron}. 
AVTacotron2~\cite{abdelaziz2021avtacotron} extends Tacotron2~\cite{shen2018tacotron2} and demonstrates improved quality than pipelined methods.

\noindent\textbf{Speech-driven facial animation.}
Early methods targeted a specific speaker or emotion~\cite{taylor2017deep,suwajanakorn2017obama}.
Recently, several methods that support an arbitrary speaker's voice~\cite{eskimez2018generating,zhou2020makeittalk,deng2021unsupervised,prajwal2020wav2lip,zhou2021pcavs,liang2022gcavt,zhang2022meta} or generate talking faces in accordance with speech emotions~\cite{deng2021unsupervised,ji2021audio} have been proposed.
Particularly, multimodal methods that consider both text and speech have improved the quality of talking face generation~\cite{yu2020multimodal,yu2021multimodal,fan2022joint}.
However, their applicability is limited because they cannot generate talking faces from only text or speech, or they have only been validated on a specific speaker.

\section{UniFLG}
\label{sec:prop}
UniFLG consists of two components: 
(1) VAE-VITS~\cite{mitsui2022endtoend}, which introduces an utterance-level latent variable into the end-to-end TTS called VITS~\cite{kim2021vits} and 
(2) a landmark decoder, which generates facial landmarks from the common representation of text and speech extracted by VAE-VITS.
Our system is trained in two stages, as depicted in \Fig{diagram}(a).
First, VAE-VITS is trained on speech and its transcriptions, and following this, the landmark decoder is trained using paired speech and facial landmarks, as well as its transcriptions with fixed VAE-VITS parameters.
UniFLG simultaneously generates speech and facial landmarks during text-driven inference, and it generates facial landmarks without using textual information during speech-driven inference, as illustrated in \Fig{diagram}(b).
We provide details regarding UniFLG in the following sections.

\subsection{VAE-VITS}
VITS models the distribution of a speech waveform $\xB$ conditioned on text $\cB$ by introducing a frame-level latent variable $\zB_f$.
The relationship between $\xB$ and $\zB_f$ is modeled by a posterior encoder and waveform decoder, and that between $\zB_f$ and $\cB$ is modeled by a prior encoder.
To model the temporal alignment between $\zB_f$ and $\cB$, they are first converted into latent representations;
$\zB_f$ is transformed into $f_\theta^\mathrm{spec}(\zB_f)$ using a normalizing flow $f_\theta$~\cite{rezende2015flow}, and $\cB$ is transformed into $\{\muB_\theta, \sigmaB_\theta\}$ using text encoder and linear projection.
Then, the monotonic alignment search (MAS)~\cite{kim2020glowtts} algorithm estimates the most probable alignment between $f_\theta^\mathrm{spec}(\zB_f)$ and $\{\muB_\theta, \sigmaB_\theta\}$.
As the alignment is not available during inference, a duration predictor is simultaneously trained; it uses the text encoder output $\hB$ to predict the duration of each phoneme $\mathbf{d}_\mathrm{spec}$.
During inference, $\{\muB_\theta, \sigmaB_\theta\}$ is expanded into frame-level using $\mathbf{d}_\mathrm{spec}$, and $f_\theta^\mathrm{spec}(\zB_f)$ is sampled from a Gaussian distibution defined by them.
Therefore, $f_\theta^\mathrm{spec}(\zB_f)$ can be regarded as a speaker-independent representation common to text and speech.

To automatically extract emotion embeddings from speech, UniFLG uses VAE-VITS, which extends VITS with an utterance encoder.
By conditioning the entire system with a one-hot speaker embedding $\gB$, the utterance encoder extracts utterance-level latent variable $\zB_u$ that represents speech emotions~\cite{mitsui2022endtoend}.

\subsection{Landmark Decoder}
The landmark decoder generates a series of facial landmarks $\YB \in \RR^{T\times N\times2}$ given $f_\theta^\mathrm{land}(\zB_f)$, a resampled version of $f_\theta^\mathrm{spec}(\zB_f)$, where $T$ and $N$ denote the number of frames and 2D keypoints, respectively.
A non-causal WaveNet~\cite{prenger2019waveglow} is used for the landmark decoder.
Following WaveNet~\cite{oord2016wavenetssw}, the emotion embedding $\zB_u$ is given as the global conditioning.

\subsubsection{Mixed-modality training}
Although $f_\theta^\mathrm{spec}(\zB_f)$ is common to text and speech, the ones that come from text and speech do not match exactly.
Therefore, the landmark decoder is trained by switching the input between text and speech at each iteration.
Given text, the duration $\mathbf{d}_\mathrm{spec}$ obtained from the MAS is multiplied by a constant to match the frame rate of $\YB$ to obtain $\mathbf{d}_\mathrm{land}$.
Thereafter, $\{\muB_\theta, \sigmaB_\theta\}$ are expanded according to $\mathbf{d}_\mathrm{land}$, and
$f_\theta^\mathrm{land}(\zB_f)$ is sampled from a Gaussian distribution defined by them.
Given speech, $f_\theta^\mathrm{spec}(\zB_f)$ extracted from $\XB_\mathrm{spec}$ is resampled by linear interpolation to obtain $f_\theta^\mathrm{land}(\zB_f)$.
The landmark decoder is trained to minimize the mean squared error between predicted and target facial landmarks.

\input{tab/objective_eval}

\subsubsection{Inference}
The flow of speech-driven inference is exactly the same as that in training.
During text-driven inference, $\mathbf{d}_\mathrm{spec}$ obtained using the duration predictor is converted into $\mathbf{d}_\mathrm{land}$, and the following flow is the same as in training.
Additionally, $\zB_u$ is extracted from speech during speech-driven inference and sampled from the prior $p(\zB_u)=\NM(\zB_u; \zeroB, \IB)$ or extracted from reference speech during text-driven inference.

\subsection{UniFLG-AS}
\label{sec:uniflg_as}
Although UniFLG eliminates speaker-dependent factors from speech using the posterior encoder and flow, it cannot generate facial landmarks from the speech of unseen speakers because these modules require speaker embeddings.
To overcome this limitation, we propose a variant of our system, UniFLG for an arbitrary speaker (UniFLG-AS), which uses a one-hot emotion embedding as $\gB$ instead of the speaker embedding, and $\zB_u$ to represent speakers instead of emotions.
The landmark decoder uses not $\zB_u$ but $\gB$ as a global condition.

\vspace{-5pt}
\section{Experiments}
\label{sec:exp}

\vspace{-1pt}
\subsection{Experimental conditions}
\vspace{-1pt}
\subsubsection{Datasets}
Although the datasets such as VoxCeleb2~\cite{chung2018voxceleb2} and MEAD~\cite{wang2020mead} have been widely used for talking face generation, they have relatively small amounts of data per speaker and no transcriptions exist.
Thus, we recorded 3,359 (1,499 normal, 860 happy, and 1,000 sad) utterances of paired speech and facial videos from a female Japanese speaker according to predefined transcripts for training the landmark decoder.
These will be hereinafter referred to as the \data{Paired} dataset.
The development and evaluation sets comprised 45 emotion-balanced utterances, respectively, and the remaining utterances were used as the training set.
For training the VAE-VITS module, we further recorded 30,542 (14,508 talk, 7,771 happy, and 8,263 sad) utterances of speech uttered by 26 speakers (eighteen females and eight males) according to predefined transcripts.
These will be hereinafter referred to as the \data{Unpaired} dataset because the facial videos are not included.
The development and evaluation sets comprised 225 speaker- and emotion-balanced utterances, respectively, and the remaining utterances were used as the training set.
All the experiments were conducted using 24~kHz/16~bit speech signals and 30 frames per second 1280$\times$720 videos.
We used 50-dimensional linguistic features for $\cB$\footnote{This is because prior research has demonstrated that incorporating accent information can enhance speech synthesis quality, particularly in pitch-accent languages such as Japanese~\cite{yasuda2019investigation}. We believe that it is sufficient to use phonemes as $\cB$ for many other languages.}, which contains phonemes, accents, and whether the current accent phrase is interrogative extracted using Open JTalk\footnote{\url{https://open-jtalk.sourceforge.net/}}.
We extracted 70 points of facial landmarks from each frame of the facial videos using OpenPose~\cite{cao2021openpose}.

\vspace{-3pt}
\subsubsection{Model architecture and training}
\vspace{-2pt}

The model architecture and training scheme of VAE-VITS were the same as those in a previous study~\cite{mitsui2022endtoend}.
For the landmark decoder, we used 16-layer non-causal WaveNet, where each layer had 192 filters with a kernel size of five and dilation factor of one, and 1$\times$1 convolution layers placed before and after the non-causal WaveNet.
A 16-dimensional $\zB_u$ was fed to each layer of the non-causal WaveNet as a global condition.
The landmark decoder was trained using an AdamW optimizer~\cite{loshchilov2019adamw} with $\beta_1=0.8, \beta_2=0.99$, and a weight decay of 0.01.
The initial learning rate was set to $2\times 10^{-4}$ and was multiplied by $0.999875$ every epoch.
The training was conducted over 10,000 iterations (approximately 150 epochs) with a batch size of 48, which took \SI{2.5}{\hour} on a single NVIDIA Tesla P40 GPU.

\vspace{-5pt}
\subsection{Results}

\input{tab/subjective_eval}
\input{tab/subjective_eval2}
\vspace{-1pt}
\subsubsection{Facial landmark prediction accuracy}
\vspace{-1pt}
\label{sec:obj_eval}

To evaluate the prediction accuracy of text- and speech-driven inference using UniFLG (hereinafter referred to as \method{UniFLG-T} and \method{UniFLG-S}, respectively), facial landmarks over the evaluation set of the \data{Paired} dataset were predicted.
For comparison, three text-driven and two speech-driven methods (presented sequentially) were used:
(1) \method{AVTacotron2}~\cite{abdelaziz2021avtacotron}, the state-of-the-art method that generates speech and facial landmarks jointly from text, similar to the proposed method,
(2) \method{TTL} (text-to-landmark) that uses the same architecture as UniFLG but trains the landmark decoder only in a text-driven manner,
(3) \method{UniFLG-P} (\method{P} stands for pipelined) that uses the proposed system, but it first synthesizes speech and then generates facial landmarks in a speech-driven manner,
(4) \method{STL} (speech-to-landmark) that uses the same architecture as UniFLG but trains the landmark decoder only in a speech-driven manner, and
(5) \method{STL-D} (\method{D} stands for direct) that does not use the posterior encoder and flow of VAE-VITS but directly feeds $\XB_\mathrm{spec}$ to the landmark decoder.
Following previous studies~\cite{zhou2020makeittalk,wang2021anyonenet}, we evaluated the accuracy of lip movements using 
the landmark distance for lips (D-LL), 
landmark velocity difference for lips (D-VL),
and difference in the open mouth area (D-A)
and the accuracy of entire face movements using 
the landmark distance (D-L) and
landmark velocity difference (D-V).
Because the sequence length of facial landmarks predicted using text-driven methods is not always the same as the target, we aligned them using dynamic time warping~\cite{berndt1994dtw}.

The obtained results are presented in \Table{obj_eval}.
Among the four text-driven methods, \method{UniFLG-T} achieved the lowest D-LL, D-A, and D-L scores.
In particular, these values were lower than those of \method{UniFLG-P}, which implies that \method{UniFLG-T} could reduce the prediction errors caused by TTS.
\method{AVTacotron2}, on the other hand, achieved the lowest D-VL and D-V scores.
This is possibly because these metrics focus on the difference between consecutive frames and \method{AVTacotron2} was the only method that generated facial landmarks autoregressively.
Among the three speech-driven methods, \method{UniFLG-S} achieved the lowest values for all evaluation metrics.
In addition, \method{UniFLG-\{T, S\}} achieved better performance than \method{TTL} and \method{STL}, respectively.
This result demonstrates the effectiveness of the proposed method trained using both text and speech.

\subsubsection{Inference speed}

Fast inference is crucial, especially for real-time applications such as human conversation and live streaming.
We measured the real time factor (RTF), the time required to generate \SI{1}{\second} speech and facial landmarks, on one NVIDIA Tesla P40 GPU.
For \method{AVTacotron2}, we included the time required for waveform generation, for which we used HiFi-GAN~\cite{kong2020hifigan} trained on all speech data in the \data{Paired} and \data{Unpaired} datasets.
The results are listed in the right-most column of \Table{obj_eval}.
As UniFLG is non-autoregressive, its RTF was generally smaller than that of the autoregressive \method{AVTacotron2}.
We also confirmed that \method{UniFLG-T} improved the RTF by 41\% over \method{UniFLG-P} because it eliminates the need for generating speech before facial landmark generation.
Among the speech-driven methods, \method{UniFLG-S} was not as fast as \method{STL-D}; however, it was still faster than any text-driven method.

\subsubsection{Generated speech and facial landmark quality}
Subjective evaluation was conducted based on the following three criteria: speech quality, facial landmark quality, and lip-sync quality\footnote{\url{https://rinnakk.github.io/research/publications/UniFLG}}.
For the second and third criteria, the facial landmarks were plotted frame by frame to construct a facial video and presented to the raters.
Four methods (\method{AVTacotron2} and \method{UniFLG-\{P, T, S\}}) were compared over the two datasets (\data{Paired}, \data{Unpaired}).
Note that the speech quality of \method{UniFLG-S} indicates the quality of the recorded speech.
Thirty-one raters participated in the evaluation, and each rater evaluated 30 samples on a five-point scale from one (bad) to five (excellent).

The obtained results are summarized in \Table{sbj_eval}.
Overall, similar trends were observed for the \data{Paired} and \data{Unpaired} datasets.
The speech quality of \method{AVTacotron2} was quite low; this is because the usage of multiple datasets, speakers, and emotions made the training difficult, resulting in the failure of alignment and stop token prediction.
\method{UniFLG-\{P, T\}} demonstrated significantly improved speech quality and those scores were close to that of \method{UniFLG-S} for the \data{Paired} dataset.
The scores slightly decreased for the \data{Unpaired} dataset, which is possibly because the amount of training data per speaker was approximately one-third of the data in the \data{Paired} dataset.
\method{UniFLG-\{P, T, S\}} achieved similar facial landmark quality and lip-sync quality scores.
These scores exceeded 4.0 and were significantly better than those of \method{AVTacotron2}.
Based on these results, we concluded that the proposed UniFLG can generate high-quality facial landmarks from either text or speech.
Furthermore, based on the fact that the scores of \method{UniFLG-T} were comparable to those of \method{UniFLG-P}, we can conclude that it can be used as a faster alternative to the pipelined counterpart.

\vspace{-2pt}
\subsubsection{Facial landmark generation for unseen speakers}
\vspace{-2pt}

To evaluate the facial landmark generation quality for unseen speakers during training, we considered 240 (80 talk, happy, and sad) utterances of speech uttered by 10 unseen speakers (five males and five females) as the \data{Unseen} dataset.
For each of the \data{Paired}, \data{Unpaired}, and \data{Unseen} datasets, we generated facial landmarks from speech using the UniFLG-AS, described in \Sec{uniflg_as} (hereinafter referred to as \method{UniFLG-AS-S}) and conducted subjective evaluation as outlined in the previous section.
For the baseline, we used \method{STL-D} described in \Sec{obj_eval}.
As \method{STL-D} does not use VAE-VITS, both the \data{Unpaired} and \data{Unseen} datasets were unseen during training, and hence, we omitted evaluation for the \data{Unpaired} dataset.

The obtained results are summarized in \Table{sbj_eval2}.
Both facial landmark and lip-sync quality of \method{UniFLG-AS-S} for the \data{Unseen} dataset outperformed those of \method{STL-D} for the \data{Paired} dataset.
This indicates that the proposed system can generate high-quality facial landmarks from the speech of unseen speakers.
Although the lip-sync quality score of \method{UniFLG-AS-S} for the \data{Unseen} dataset exceeded 4.0, it was slightly worse than that for the \data{Unpaired} dataset.
The number of speakers seen while training VAE-VITS was 27 in total, which we assume was insufficient to represent all unseen speakers with the learned latent space.
Narrowing this performance gap by using more speakers for training VAE-VITS will form a part of future research.

\vspace{-4pt}
\section{Conclusions}
\vspace{-2pt}
\label{sec:conlcusion}

This paper proposed UniFLG, which integrates audiovisual speech synthesis and speech-driven facial animation frameworks.
The involved experimental evaluation demonstrated that the proposed system could generate higher quality facial landmarks than conventional methods from either text or speech.
Moreover, UniFLG-AS, a variant of UniFLG, could generate natural facial landmarks even from the speech of unseen speakers.
Future research would involve attempts to extend the proposed system to an end-to-end framework that includes the training of a video generation network.
It would also be advantageous to integrate UniFLG and UniFLG-AS to generate facial landmarks from the speech of arbitrary speakers and emotions.

\bibliographystyle{IEEEbib}
\bibliography{ref/speech, ref/image, ref/ml}

\end{document}

%% file: math.tex
\newcommand{\argmax}{\mathop{\mathrm{argmax}}\limits}
\newcommand{\argmin}{\mathop{\mathrm{argmin}}\limits}
\newcommand\diag{\mathop{\mathrm{diag}}}

\newcommand{\epsilonB}{\bm{\epsilon}}
\newcommand{\muB}{\bm{\mu}}
\newcommand{\sigmaB}{\bm{\sigma}}
\newcommand{\SigmaB}{\bm{\Sigma}}
\newcommand{\LambdaB}{\bm{\Lambda}}
\newcommand{\alphaB}{\bm{\alpha}}
\newcommand{\betaB}{\bm{\beta}}
\newcommand{\thetaB}{\bm{\theta}}
\newcommand{\ThetaB}{\bm{\Theta}}
\newcommand{\XiB}{\bm{\Xi}}
\newcommand{\UpsilonB}{\bm{\Upsilon}}
\newcommand{\piB}{\bm{\pi}}
\newcommand{\OmegaB}{\bm{\Omega}}
\newcommand{\PsiB}{\bm{\Psi}}
\newcommand{\etaB}{\bm{\eta}}
\newcommand{\DeltaB}{\bm{\Delta}}
\newcommand{\PhiB}{\bm{\Phi}}
\newcommand{\phiB}{\bm{\phi}}

\newcommand{\AB}{\mathbf{A}}
\newcommand{\BB}{\mathbf{B}}
\newcommand{\bB}{\mathbf{b}}
\newcommand{\cB}{\mathbf{c}}
\newcommand{\CB}{\mathbf{C}}
\newcommand{\DB}{\mathbf{D}}
\newcommand{\fB}{\mathbf{f}}
\newcommand{\FB}{\mathbf{F}}
\newcommand{\gB}{\mathbf{g}}
\newcommand{\hB}{\mathbf{h}}
\newcommand{\HB}{\mathbf{H}}
\newcommand{\IB}{\mathbf{I}}
\newcommand{\kB}{\mathbf{k}}
\newcommand{\KB}{\mathbf{K}}
\newcommand{\LB}{\mathbf{L}}
\newcommand{\MB}{\mathbf{M}}
\newcommand{\mB}{\mathbf{m}}
\newcommand{\OB}{\mathbf{O}}
\newcommand{\PB}{\mathbf{P}}
\newcommand{\pB}{\mathbf{p}}
\newcommand{\QB}{\mathbf{Q}}
\newcommand{\RB}{\mathbf{R}}
\newcommand{\rB}{\mathbf{r}}
\newcommand{\SB}{\mathbf{S}}
\newcommand{\TB}{\mathbf{T}}
\newcommand{\UB}{\mathbf{U}}
\newcommand{\uB}{\mathbf{u}}
\newcommand{\vB}{\mathbf{v}}
\newcommand{\VB}{\mathbf{V}}
\newcommand{\wB}{\mathbf{w}}
\newcommand{\WB}{\mathbf{W}}
\newcommand{\xB}{\mathbf{x}}
\newcommand{\XB}{\mathbf{X}}
\newcommand{\yB}{\mathbf{y}}
\newcommand{\YB}{\mathbf{Y}}
\newcommand{\zB}{\mathbf{z}}
\newcommand{\ZB}{\mathbf{Z}}
\newcommand{\zeroB}{\mathbf{0}}
\newcommand{\oneB}{\mathbf{1}}

\newcommand{\FM}{\mathcal{F}}
\newcommand{\KM}{\mathcal{K}}
\newcommand{\LM}{\mathcal{L}}
\newcommand{\NM}{\mathcal{N}}
\newcommand{\OM}{\mathcal{O}}
\newcommand{\RM}{\mathcal{R}}
\newcommand{\SM}{\mathcal{S}}
\newcommand{\UM}{\mathcal{U}}

\newcommand{\EE}{\mathbb{E}}
\newcommand{\RR}{\mathbb{R}}
\newcommand{\KL}{\mathrm{KL}}

\newcommand{\drm}{{\mathrm{d}}}
\newcommand{\frm}{{\mathrm{f}}}

\newcommand{\fzero}{{$f_\mathrm{o}~$}}
\newcommand{\fzeronosp}{{$f_\mathrm{o}$}}

%% file: tab/objective_eval.tex
\begin{table*}[t]
\small
\caption{Objective evaluation of text-driven and speech-driven facial landmark generation in terms of the lip landmark prediction accuracy (D-LL, D-VL, D-A), entire landmark prediction accuracy (D-L, D-V), and real time factor (RTF) of inference.}
\vspace{-10pt}
\renewcommand{\arraystretch}{0.9}
\label{tbl:obj_eval}
\begin{center}
\begin{tabular}{l l ccc cc c}\toprule
\textbf{Input} & \textbf{Method} & \textbf{D-LL} $\downarrow$ [\%] & \textbf{D-VL} $\downarrow$ [\%] & \textbf{D-A} $\downarrow$ [\%] & \textbf{D-L} $\downarrow$ [\%] & \textbf{D-V} $\downarrow$ [\%] & \textbf{RTF} $\downarrow$ \\
\midrule
\multirow{4}{*}{Text} 
& \method{AVTacotron2}~\cite{abdelaziz2021avtacotron} & 10.2 & \textbf{1.65} & 18.1 & 4.50 & \textbf{0.572} & 0.088 \\
& \method{TTL} & 9.16 & 1.80 & 16.7 & 3.91 & 0.633 & 0.027 \\
& \method{UniFLG-P} & 8.69 & 1.83 & 16.5 & 3.70 & 0.653 & 0.038 \\
& \method{UniFLG-T} & \textbf{8.41} & 1.86 & \textbf{16.2} & \textbf{3.55} & 0.660 & 0.027 \\
\midrule
\multirow{3}{*}{Speech} 
& \method{STL} & 9.00 & 1.86 & 11.5 & 3.83 & 0.721 & 0.014\\
& \method{STL-D} & 9.59 & 2.08 & 12.1 & 4.11 & 0.803 & 0.004\\
& \method{UniFLG-S} & \textbf{8.82} & \textbf{1.66} & \textbf{11.1} & \textbf{3.76} & \textbf{0.629} & 0.014\\
\bottomrule
\end{tabular}
\vspace{-10pt}
\end{center}
\end{table*}

%% file: tab/subjective_eval.tex
\begin{table*}[t]
\small
\caption{Subjective evaluation of text-driven and speech-driven systems in terms of the speech, facial landmark, and lip-sync quality.}
\vspace{-15pt}
\renewcommand{\arraystretch}{0.9}
\label{tbl:sbj_eval}
\begin{center}
\begin{tabular}{l l ccc}\toprule
\textbf{Data} & \textbf{Method} & \textbf{Speech} $\uparrow$ & \textbf{Landmark} $\uparrow$ & \textbf{Lip-Sync} $\uparrow$ \\
\midrule
\multirow{4}{*}{\data{Paired}}
& \method{AVTacotron2} & 1.85$\pm$0.20 & 3.71$\pm$0.22 & 3.79$\pm$0.24 \\
& \method{UniFLG-P} & 4.30$\pm$0.16 & 4.26$\pm$0.15 & 4.26$\pm$0.16 \\
& \method{UniFLG-T} & 4.24$\pm$0.15 & \textbf{4.28$\pm$0.15} & \textbf{4.31$\pm$0.14} \\
& \method{UniFLG-S} & \textbf{4.57$\pm$0.11} & 4.20$\pm$0.15 & 4.20$\pm$0.15 \\
\midrule
\multirow{4}{*}{\data{Unpaired}}
& \method{AVTacotron2} & 1.99$\pm$0.16 & 3.52$\pm$0.17 & 3.66$\pm$0.19 \\
& \method{UniFLG-P} & 3.85$\pm$0.20 & \textbf{4.12$\pm$0.17} & 4.12$\pm$0.19 \\
& \method{UniFLG-T} & 3.88$\pm$0.16 & \textbf{4.12$\pm$0.15} & \textbf{4.31$\pm$0.15} \\
& \method{UniFLG-S} & \textbf{4.43$\pm$0.16} & 4.08$\pm$0.17 & 4.18$\pm$0.18 \\
\bottomrule
\end{tabular}
\vspace{-15pt}
\end{center}
\end{table*}

%% file: tab/subjective_eval2.tex
\begin{table}[t]
\small
\caption{Subjective evaluation of speech-driven facial landmark generation for the \data{Paired}, \data{Unpaired}, and \data{Unseen} datasets.}
\vspace{-15pt}
\renewcommand{\arraystretch}{0.9}
\label{tbl:sbj_eval2}
\begin{center}
\begin{tabular}{l l cc}\toprule
\textbf{Data} & \textbf{Method} & \textbf{Landmark} $\uparrow$ & \textbf{Lip-Sync} $\uparrow$ \\
\midrule
\multirow{2}{*}{\data{Paired}}
& \method{STL-D} & 3.95$\pm$0.16 & 4.01$\pm$0.16 \\
& \method{UniFLG-AS-S} & \textbf{4.45$\pm$0.12} & \textbf{4.44$\pm$0.12} \\
\midrule
\data{Unpaired}
& \method{UniFLG-AS-S} & 4.23$\pm$0.14 & 4.22$\pm$0.14 \\
\midrule
\multirow{2}{*}{\data{Unseen}}
& \method{STL-D} & 3.96$\pm$0.17 & 3.68$\pm$0.18 \\
& \method{UniFLG-AS-S} & \textbf{4.26$\pm$0.13} & \textbf{4.04$\pm$0.17} \\
\bottomrule
\end{tabular}
\vspace{-20pt}
\end{center}
\end{table}